# Spatial features of $CO_2$ for occupancy detection in a naturally ventilated school building

Qirui Huang[*], Marc Syndicus, Jérôme Frisch, Christoph van Treeck

*Institute of Energy Efficiency and Sustainable Building (E3D), RWTH Aachen University, Aachen, Germany*

## ARTICLE INFO

## ABSTRACT

Accurate occupancy information helps to improve building energy efficiency and occupant comfort. Occupancy detection methods based on $CO_2$ sensors have received attention due to their low cost and low intrusiveness. In naturally ventilated buildings, the accuracy of $CO_2$-based occupancy detection is generally low in related studies due to the complex ventilation behavior and the difficulty in measuring the actual air exchange through windows. In this study, we present two novel features for occupancy detection based on the spatial distribution of the $CO_2$ concentration. After a quantitative analysis with Support Vector Machine (SVM) as classifier, it was found that the accuracy of occupancy state detection in naturally ventilated rooms could be improved by up to 14.8 percentage points compared to the baseline, reaching 83.2 % (F1 score 0.84) without any ventilation information. With ventilation information, the accuracy reached 87.6 % (F1 score 0.89). The performance of occupancy quantity detection was significantly improved by up to 25.3 percentage points versus baseline, reaching 56 %, with root mean square error (RMSE) of 11.44 occupants, using only $CO_2$-related features. Additional ventilation information further enhanced the performance to 61.8 % (RMSE 9.02 occupants). By incorporating spatial features, the model using only $CO_2$-related features revealed similar performance as the model containing additional ventilation information, resulting in a better low-cost occupancy detection method for naturally ventilated buildings.

## 1. Introduction

Detailed and accurate occupancy information is an integral part of modern building management. Related studies [1,2] have demonstrated the significance of occupancy information for reducing building energy consumption (with potential savings of up to 30 %) and ensuring occupant comfort. Occupancy detection technologies based on $CO_2$ sensors have gained attention in recent years, mostly due to their low cost and non-invasiveness [3]. Non-invasiveness includes not only technical data security but also users' subjective willingness. Additionally, the COVID-19 pandemic has brought $CO_2$ to the forefront as an indicator of indoor air quality, leading to widespread availability and use of low-cost $CO_2$ measurement devices [4–9]. Using these existing $CO_2$ sensors, originally intended for monitoring air quality, to detect occupancy information means no additional hardware expenditure for users.

According to a review by Rueda et al. [3], $CO_2$-based occupancy detection methods have been extensively studied. Relevant studies with a high performance of $CO_2$-based methods that have been validated in non-laboratory settings are summarized in Table 1. The average performance in Table 1 is the average of the accuracy and RMSE over all rooms in each study. However, the studies reviewed by Rueda et al. [3] have primarily focused on occupancy detection in buildings with mechanical ventilation and presumably steady-state air exchange rates. The detection performance in purely naturally ventilated buildings has been rarely studied. As ventilation in naturally ventilated buildings is a decision made by the occupants and a spontaneous activity difficult to predict, obtaining accurate occupancy information becomes even more challenging and crucial for the building management system, influencing decisions related to heating, lighting, and other aspects [2,10–12].

In the study of Calí et al. [14], the performance of occupancy detection in naturally ventilated buildings demonstrated lower efficacy. This can be attributed, firstly, to the inherent challenge of measuring the air exchange rate in such buildings. Secondly, the heterogeneous $CO_2$ distribution indoors introduces another complication, where the sensor's location significantly influences both the measurement results and prediction performance [17–22].

The vertical gradient of $CO_2$ due to rising thermal plumes of occupants in rooms with mechanical ventilation has been reported in several

**Table 1**

Overview of $CO_2$-based occupancy detection state [+] and quantity [#] performance in mechanically and naturally ventilated buildings in relevant studies.

| Source | Method | Average performance (mech. vent.) | Average performance (nat. vent.) |
|--------|--------|-----------------------------------|----------------------------------|
| [13][1] | data-driven | – | [+]79.07 % |
| [14][2] | physical model | [+]88.25 % | [+]80.13 % |
| | | [#]80.13 % | [#]65.90 % |
| [14][3] | physical model | [+]89.95 % | [+]84.53 % |
| | | [#]72.85 % | [#]64.93 % |
| [15] | physical model | [#]RMSE 12.8[*] | – |
| | data-driven | [#]76 % | – |
| | | [#]RMSE 12.1[*] | – |
| [16] | data-driven | [#]75 % | – |

[1] Average performance of models using only $CO_2$ as parameter.
[2] Performance of scenario 3 without window opening information.
[3] Performance of scenario 2 with window opening information.
[*] Unit of RMSE: occupancy counts.

related studies [18–21,23,24]. The vertical gradient is defined as the difference or stratification of $CO_2$ concentration at different heights in the room, excluding cases of mixed ventilation. This phenomenon was also observed in the study of Mahyuddin et al. [21] and our previous field study [22], both conducted in a school with natural ventilation. Furthermore, Choi et al. [17] reported a heterogeneous $CO_2$ distribution in the horizontal plane, influenced by occupancy and ventilation behavior. These spatial distribution properties of $CO_2$ have not been utilized in occupancy detection studies. Typically, the spatial layout of the sensors has been mostly considered as a means to validate the algorithm's robustness.

In this paper, we introduce two novel features derived from the spatial distribution of $CO_2$ to enhance the occupancy detection performance in naturally ventilated buildings.

## 2. Methods

### 2.1. Data collection and preprocessing

The data utilized in this study were gathered through a field study conducted in a public school in Düren, Germany (Köppen-Geiger classification Cfb [25]), spanning from December 8th to 21st, 2021, and May 18th to June 14th, 2022 [5]. In this field study, multiple measurement devices were installed in three classrooms. Each device was equipped with two calibrated $CO_2$ sensors (Sensirion® SCD30) to collect data at 15-second intervals, excluding nighttime and weekends [22].

According to the manufacturer's datasheet [26], the measurement accuracy of each $CO_2$ sensor is within ± 30 ppm. By comparing the data from the two sensors in each measurement device, the inter-sensor variance was consistent with the manufacturer's statement [22]. To minimize the inter-sensor error, we used the average of the two $CO_2$ sensors in each measurement device to represent the local $CO_2$ level. Missing values were filled in using the nearest neighbor method. Fig. 1 illustrates the installation locations of the sensors used in this study. The measurement devices with a height of 0.6 m were mounted on the undersurface of the desk, while other devices were mounted on cabinets or walls.

All three classrooms feature windows that can be opened on one of the walls for ventilation. Windows that cannot be opened are not depicted in Fig. 1 as they do not impact ventilation. The classrooms' specific dimensions and the openable windows' area are detailed in Table 2.

Room occupancy and window openings were primarily gathered through questionnaires. Teachers documented the number of students in each lesson, described the natural ventilation regime performed in that lesson, and could remark special events that may have happened (e.g., noise from the outside which lead to window closure). During the summer, we enhanced the resolution of window opening data by installing reed switches on the windows. Through the evaluation of both questionnaires and data from reed switches, the ventilation rating, based on the window opening status was normalized to a range of 0–1, zero indicates that all windows were closed, while one signifies that all windows were open.

To mitigate the influence of minor fluctuations in the sensor data and to facilitate better comparison with the results from Zuraimi et al. [15], the data were normalized and then aggregated into 5-minute intervals. Details regarding the sample size of the aggregated data and the occupancy of each room can be found in Table 3.

### 2.2. Feature selection

The primary objectives of this study are to compare the performance differences in occupancy detection arising from various $CO_2$-related features and to identify which features offer crucial information for detecting both occupancy state and quantity in naturally ventilated rooms. Numerous studies have demonstrated a significant correlation between $CO_2$ concentration and its temporal features, such as the first-order difference and moving average of $CO_2$ measurements, with occupancy state and quantity [3,14–16,27,28–30]. Zuraimi et al. [15] found that using the $CO_2$ average and its first-order difference yields the best performance.

Hence, we adopted the combination of average and first-order differential $CO_2$ concentration as the baseline for evaluation. Additionally, we selected the two measurement devices in each room that were horizontally closest, situated at a similar distance from the windows, and had a height difference greater than 1 m. We used these pairs to calculate the vertical difference in $CO_2$ concentration, representing the vertical gradient phenomenon. For the horizontal difference, the two devices closest and farthest from the windows with the smallest height difference were chosen to represent the $CO_2$ changes due to ventilation. For room 1, the horizontal difference was not utilized as a feature due to the malfunctioning of the measurement device adjacent to the windows. The selected features are outlined in Table 4. Based on these criteria, we compared the occupancy detection differences for various feature combinations as part of our evaluation.

### 2.3. Model training

In their comprehensive review paper, Rueda et al. [3] noted that Support Vector Machines (SVMs) have been widely employed for occupancy detection in relevant studies due to their high performance. SVMs exhibit superior accuracy in detecting occupancy quantity compared to other physical model-based and data-driven methods, achieving up to 76 % accuracy, while using the same dataset and utilizing only $CO_2$-related features [3,15].

Therefore, for a quantitative comparison of various feature combinations, we used a SVM with a radial basis function (RBF) kernel as the classifier, as described in Equation (1). To ensure that the optimal values of the RBF kernel parameter γ and the penalty parameter of error term $C$ are selected, we used grid search with $k$-fold cross-validation where $k$ = 5 after data splitting. The grid search was performed in the range of $2^{-10}$ to $2^{10}$ for $C$ and 0–$2^{10}$ for γ. If the optimal solution appears at the grid boundary, then the grid is expanded for further search. The above procedure was implemented in Python using the libsvm-based Python library scikit-learn [31].

$$K(x_i, x_j) = exp(-\gamma \|x_i - x_j\|^2), \quad \gamma > 0 \tag{1}$$

We used 60 % of the data for training and 40 % for evaluation. Three rounds of data splitting were performed for each room to obtain lower bias results, and the average of the three rounds was used in the final evaluation. The data used for training contained data from both summer and winter to capture as many occupancy features in different situations as possible. The entire process is visualized in Fig. 2.

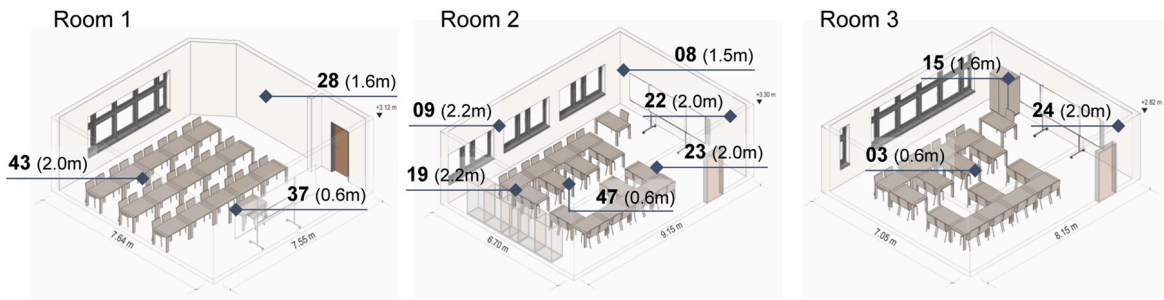

**Fig. 1.** Installation locations of measurement devices in three classrooms. The devices that were not used due to malfunction are not included. Numbers indicate the reference number of the randomly selected and installed devices; numbers in parentheses indicate the height installed.

**Table 2**
Dimensions and openable window areas of the classrooms in Fig. 1.

| Room | Length | Width | Height | Window area |
|---|---|---|---|---|
| 1 | 7.64 m | 7.55 m | 3.12 m | 7.956 m² |
| 2 | 9.15 m | 6.70 m | 3.30 m | 8.400 m² |
| 3 | 8.15 m | 7.05 m | 2.82 m | 9.972 m² |

Since the rooms were unoccupied for the majority of time, balanced class weights and accuracy scores were applied during cross-validation and training to avoid inflated performance estimates for the unbalanced dataset [31,32].

Given that the RBF kernel is nonlinear, the contribution of features to the model is not directly obtainable. Therefore, we used the permutation importance (number of times to permute a feature = 5) for feature evaluation. This method evaluates the importance of features by randomly disrupting individual features based on the variation in performance [31]. A higher importance indicates that the model relies more on the feature.

### 2.4. Metrics

In the quantitative evaluation, we used the accuracy score (Equation (2)) and the F1 score (Equation (3) - (5)) to evaluate the performance of the occupancy state classifier. The accuracy score is obtained when the predicted value ($\hat{y}_i$) strictly matches the corresponding ground truth label ($y_i$) where $1(x)$ is the indicator function. If the entire set of predicted values exactly matches the ground truth, then the subset accuracy score is 1; otherwise, it is 0. The F1 score is the harmonic mean of the precision and recall. Both scores reach their best value at one and worst at zero.

$$\text{accuracy}(y, \hat{y}) = \frac{1}{n_{\text{samples}}} \sum_{i=0}^{n_{\text{samples}}-1} 1(\hat{y}_i = y_i) \quad (2)$$

$$\text{F1}(y, \hat{y}) = \frac{\text{Precision} \cdot \text{Recall}}{\text{Precision} + \text{Recall}} \quad (3)$$

where:

$$\text{Precision} = \frac{\text{True Positives}}{\text{True Positives} + \text{False Positives}} \quad (4)$$

$$\text{Recall} = \frac{\text{True Positives}}{\text{True Positives} + \text{False Negatives}} \quad (5)$$

The root mean square error (RMSE, see equation (6)) was used to evaluate the occupancy quantity classifiers for measuring the difference between the ground truth label and the predicted quantity. The unit of RMSE in this paper is the occupancy count (shortened to occupants).

$$\text{RMSE}(y, \hat{y}) = \sqrt{\text{MSE}(y, \hat{y})} = \sqrt{\frac{1}{n_{\text{samples}}} \sum_{i=0}^{n_{\text{samples}}-1} (y_i - \hat{y}_i)^2} \quad (6)$$

## 3. Results

### 3.1. Correlation between $CO_2$ and occupancy

The relationship between $CO_2$ concentration and the number of occupants in each classroom is shown in Fig. 3, where the blue and red points indicate sensor data in the winter and summer period, respectively. No significant correlation between them can be observed from the distribution of data points, precisely due to the complex ventilation behavior in naturally ventilated buildings. When occupied, the $CO_2$ concentration may remain low with proper ventilation. It could persist at high levels while unoccupied due to the carry-over effect reported in our previous study [5]. The $CO_2$ concentration remained high after occupancy due to the airtightness, resulting in only a slight decrease, leading to a temporal profile similar to that observed when the room was occupied. Despite variations in occupancy, notable differences in $CO_2$ concentration were not evident between winter and summer.

To quantify the relationship between $CO_2$-related features and occupancy, we used the Spearman Rank Order Correlation Coefficient (SROCC, $r_s$) [33] to assess the correlation between them. The SROCC served to assess the statistical dependence between the rankings of the two variables, and the SROCC evaluates monotonic relationships, whether or not they are linear. The SROCCs between $CO_2$ features and occupancy for each room in Table 5 show that the correlation between $CO_2$ concentration and the number of occupants in the case of natural ventilation is not significant due to the complex ventilation behavior. In contrast to the $CO_2$ data, the physical features based on the spatial distribution of $CO_2$, especially the vertical differences of $CO_2$, correlate significantly better with occupancy information compared to the former.

**Table 3**
Sample size of aggregated data and occupancy in each room.

| Room | Sample size (total) | Sample size (occupied) | Percentage of OCC | Average of OCC |
|---|---|---|---|---|
| 1 | 4341 | 664 | 15.30% | 12 p. |
| 2 | 4324 | 1044 | 24.14% | 24 p. |
| 3 | 4334 | 842 | 19.43% | 21 p. |

Note. OCC: occupancy counts, p.: persons

**Table 4**
Description of selected features.

| Feature | Description | Formula |
|---|---|---|
| AVG | average of $CO_2$ in each room | $\overline{CO}_2(t)$ |
| FD | first-order difference of AVG | $\overline{CO}_2(t = i) - \overline{CO}_2(t = i - 1)$ |
| VD | vertical difference of $CO_2$ | room 1: $CO_2(t, x = 43) - CO_2(t, x = 37)$ |
|  |  | room 2: $CO_2(t, x = 19) - CO_2(t, x = 47)$ |
|  |  | room 3: $CO_2(t, x = 24) - CO_2(t, x = 03)$ |
| FDVD | first-order difference of VD | $VD(t = i) - VD(t = i - 1)$ |
| HD | horizontal difference of $CO_2$ | room 1: n.a. |
|  |  | room 2: $CO_2(t, x = 19) - CO_2(t, x = 09)$ |
|  |  | room 3: $CO_2(t, x = 24) - CO_2(t, x = 15)$ |
| VENT | ventilation rating | 0–1, 0: all windows closed, 1: all windows opened |

Note. x: the reference number of the measurement devices in Fig. 1, t: time. n.a.: not available due to malfunction of the measurement device adjacent to windows.

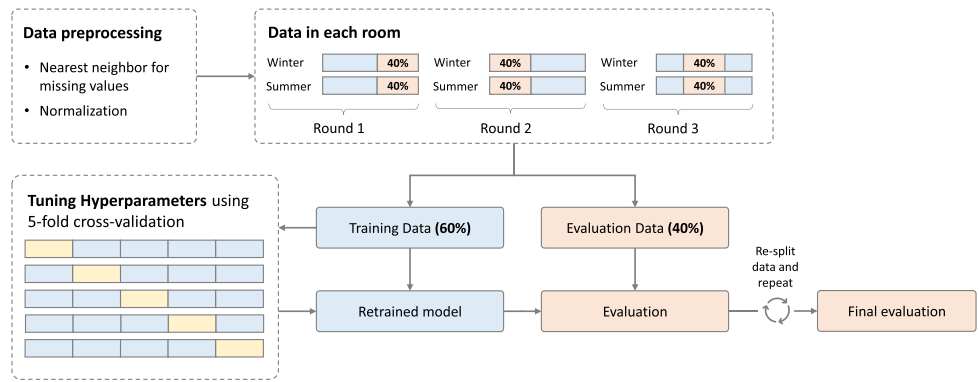

**Fig. 2.** Flow chart of data preprocessing, splitting, and model training. The dataset was split into three rounds, each containing data from both summer and winter.

The correlation between the vertical differences of $CO_2$ and occupancy can also be observed in Fig. 4, where the red bars represent the occupancy quantity and the black lines indicate the temporal changes of $CO_2$ vertical differences. Since the relationship between occupancy and $CO_2$ vertical differences behaves similarly across all three classrooms, we visualized the data for room 1 as an example here.

### 3.2. Occupancy state detection

The performance of the occupancy state classifiers, using different feature combinations in each room as well as their average values, is presented in Table 6.

The features utilized are categorized into three groups: 1) temporal features (AVG and FD) serving as a baseline for comparison, 2) spatial features (VD and HD), and 3) combined features that integrate the aforementioned features along with the first-order differentiation of VD. These features are organized in ascending order based on their average accuracy within each category.

**Table 5**
Spearman rank order correlation coefficients (SROCC) $r_s$ between $CO_2$ features and occupancy in each room.

| Room | $r_s$(AVG, OCC) | $r_s$(VD, OCC) | $r_s$(HD, OCC) |
|---|---|---|---|
| 1 | 0.50 | 0.71 | n.a. |
| 2 | 0.38 | 0.67 | 0.60 |
| 3 | 0.57 | 0.73 | 0.56 |

Note. OCC: occupancy counts. n.a.: not available due to malfunction of the measurement device adjacent to windows in room 1. For an explanation of feature abbreviations, please refer to Table 4.

The accuracy and the F1 score (average over three rounds, see Table 6) of room 1 are relatively lower compared to the other two rooms. This could be attributed to room 1 having the lowest occupancy ratio (see Table 3), with only 15.30 % of the time being occupied.

Fig. 5 illustrates the performance comparison of the occupancy state classifiers using the average accuracy and the F1 score. In comparison to the baseline, the utilization of spatial features based on the vertical difference of $CO_2$ resulted in a notable improvement, increasing the accuracy by 10.88 percentage points to 79.22 %. Further enhancement was achieved by incorporating horizontal differences, leading to a performance improvement of 13.83 percentage points, reaching 82.17 %.

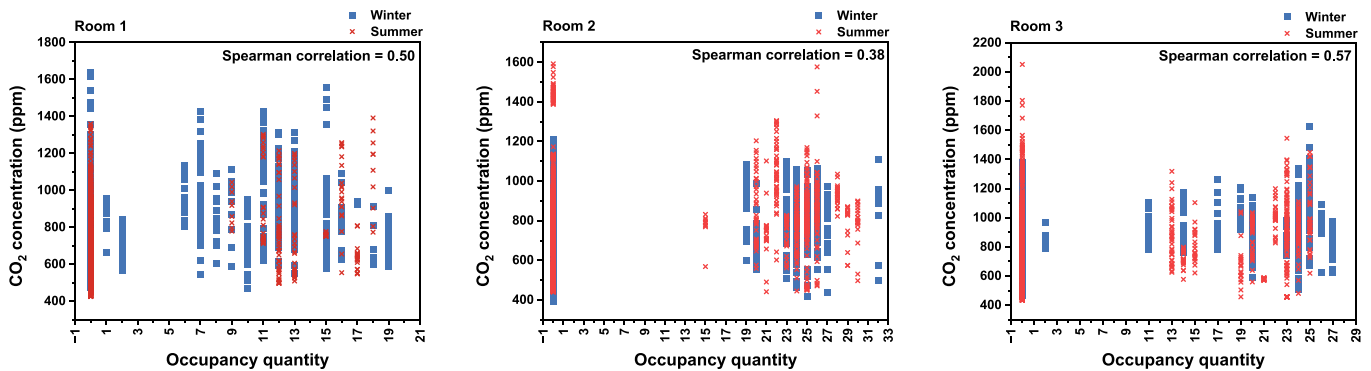

**Fig. 3.** The relationship between $CO_2$ concentration and occupancy quantity in each classroom. Blue and red points represent data collected in winter and summer, respectively.

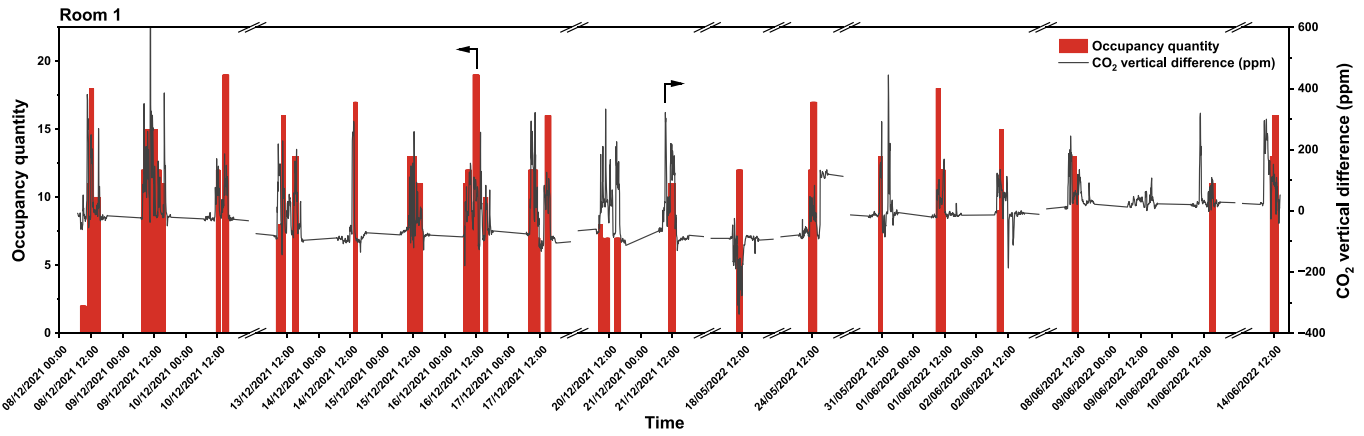

**Fig. 4.** Temporal changes in occupancy and vertical differences of $CO_2$ in room 1 as an example. Red bars represent the number of occupants and black lines the temporal changes in $CO_2$ vertical differences. Longer unoccupied periods are hidden to improve readability.

The performance could be further improved by combining temporal and spatial features. However, the average accuracy of the two best-performing scenarios (AVG + FD + VD + HD and AVG + FD + VD + FDVD + HD) might be slightly overestimated due to the absence of data using the horizontal difference feature in room 1. Despite this limitation, the combined features significantly enhanced the performance of the occupancy state classifiers compared to the baseline.

We selected the best scenarios for each room among the three categories and across three different days for qualitative analysis of the

**Table 6**
Performance of occupancy state classifiers (average and standard deviation of three rounds) using only $CO_2$-related features.

| Feature | Room | Accuracy* | Average Accuracy* | Precision* | Recall* | F1-Score* | Average F1-Score* |
|---|---|---|---|---|---|---|---|
| **Temporal Features (baseline):** | | | | | | | |
| AVG + FD | 1 | 70.67 % (0.17) | 68.34 % (0.03) | 0.86 (0.03) | 0.71 (0.17) | 0.73 (0.13) | 0.71 (0.03) |
| | 2 | 63.67 % (0.08) | | 0.78 (0.05) | 0.64 (0.08) | 0.66 (0.07) | |
| | 3 | 70.67 % (0.03) | | 0.80 (0.01) | 0.71 (0.03) | 0.73 (0.03) | |
| **Spatial Features:** | | | | | | | |
| VD | 1 | 75.00 % (0.14) | 79.22 % (0.03) | 0.86 (0.02) | 0.75 (0.14) | 0.78 (0.10) | 0.81 (0.02) |
| | 2 | 82.33 % (0.04) | | 0.88 (0.02) | 0.82 (0.04) | 0.84 (0.04) | |
| | 3 | 80.33 % (0.08) | | 0.85 (0.04) | 0.80 (0.08) | 0.82 (0.07) | |
| VD + HD | 1 | n.a. | 82.17 % (0.01) | n.a. | n.a. | n.a. | 0.84 (0.01) |
| | 2 | 81.67 % (0.04) | | 0.88 (0.01) | 0.82 (0.04) | 0.83 (0.04) | |
| | 3 | 82.67 % (0.05) | | 0.85 (0.03) | 0.83 (0.05) | 0.84 (0.04) | |
| **Combined Features:** | | | | | | | |
| AVG + VD + FDVD | 1 | 72.00 % (0.09) | 78.33 % (0.04) | 0.86 (0.02) | 0.72 (0.09) | 0.76 (0.07) | 0.81 (0.03) |
| | 2 | 81.33 % (0.05) | | 0.88 (0.02) | 0.81 (0.05) | 0.83 (0.05) | |
| | 3 | 81.67 % (0.05) | | 0.86 (0.03) | 0.82 (0.05) | 0.83 (0.04) | |
| FD + VD | 1 | 75.00 % (0.14) | 78.78 % (0.03) | 0.89 (0.02) | 0.75 (0.15) | 0.78 (0.12) | 0.81 (0.02) |
| | 2 | 82.67 % (0.04) | | 0.88 (0.02) | 0.82 (0.04) | 0.84 (0.04) | |
| | 3 | 78.67 % (0.06) | | 0.86 (0.03) | 0.79 (0.06) | 0.81 (0.05) | |
| VD + FDVD | 1 | 75.00 % (0.14) | 78.89 % (0.03) | 0.87 (0.02) | 0.75 (0.14) | 0.77 (0.11) | 0.80 (0.02) |
| | 2 | 82.00 % (0.04) | | 0.88 (0.02) | 0.82 (0.04) | 0.83 (0.04) | |
| | 3 | 79.67 % (0.07) | | 0.86 (0.04) | 0.80 (0.07) | 0.81 (0.06) | |
| AVG + FD + VD + FDVD | 1 | 73.67 % (0.08) | 79.56 % (0.04) | 0.88 (0.02) | 0.74 (0.08) | 0.77 (0.06) | 0.82 (0.03) |
| | 2 | 83.00 % (0.02) | | 0.88 (0.01) | 0.83 (0.02) | 0.85 (0.02) | |
| | 3 | 82.00 % (0.04) | | 0.87 (0.03) | 0.82 (0.04) | 0.83 (0.04) | |
| AVG + VD | 1 | 74.67 % (0.11) | 80.11 % (0.04) | 0.87 (0.02) | 0.75 (0.11) | 0.78 (0.08) | 0.82 (0.03) |
| | 2 | 82.67 % (0.03) | | 0.87 (0.02) | 0.83 (0.03) | 0.84 (0.03) | |
| | 3 | 83.00 % (0.05) | | 0.85 (0.03) | 0.83 (0.05) | 0.84 (0.04) | |
| FD + VD + HD | 1 | n.a. | 80.33 % (0.02) | n.a. | n.a. | n.a. | 0.82 (0.02) |
| | 2 | 82.33 % (0.05) | | 0.88 (0.02) | 0.82 (0.05) | 0.84 (0.05) | |
| | 3 | 78.33 % (0.04) | | 0.86 (0.02) | 0.78 (0.08) | 0.80 (0.08) | |
| AVG + FD + VD | 1 | 76.00 % (0.11) | 80.67 % (0.03) | 0.88 (0.02) | 0.76 (0.11) | 0.79 (0.09) | 0.82 (0.02) |
| | 2 | 83.00 % (0.04) | | 0.87 (0.02) | 0.83 (0.04) | 0.84 (0.03) | |
| | 3 | 83.00 % (0.04) | | 0.86 (0.02) | 0.83 (0.04) | 0.84 (0.04) | |
| AVG + FD + VD + HD | 1 | n.a. | 82.67 % (0.01) | n.a. | n.a. | n.a. | 0.84 (0.02) |
| | 2 | 84.00 % (0.02) | | 0.87 (0.02) | 0.84 (0.02) | 0.85 (0.03) | |
| | 3 | 81.33 % (0.05) | | 0.85 (0.03) | 0.81 (0.05) | 0.82 (0.04) | |
| AVG + FD + VD + FDVD + HD | 1 | n.a. | 83.17 % (0.01) | n.a. | n.a. | n.a. | 0.84 (0.01) |
| | 2 | 82.00 % (0.03) | | 0.88 (0.02) | 0.82 (0.04) | 0.83 (0.04) | |
| | 3 | 84.33 % (0.03) | | 0.86 (0.02) | 0.84 (0.03) | 0.85 (0.03) | |

* Values in parentheses indicate standard deviations. n.a.: not available due to malfunction of the measurement device adjacent to windows in room 1. Results are listed in ascending order of average accuracy in each category. For an explanation of feature abbreviations, please refer to Table 4.

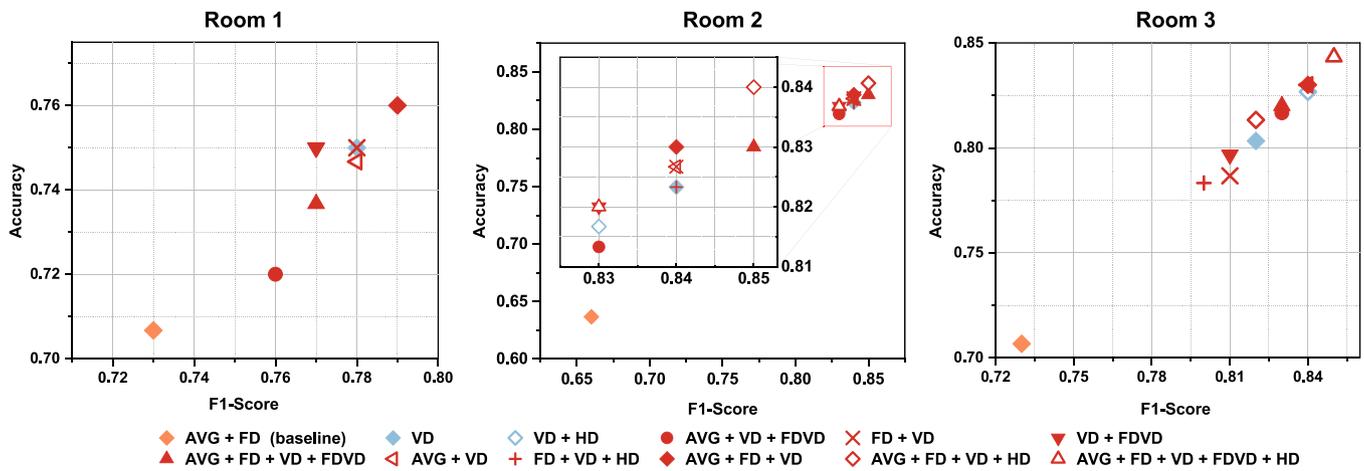

**Fig. 5.** Visualization of quantitative comparisons of state classifier performance. Orange points represent classifier performance using temporal features as the baseline, blue for spatial features, and red for combined features. For an explanation of feature abbreviations, please refer to Table 4.

prediction results, as depicted in Fig. 6. The black lines represent the room's average $CO_2$ concentration, and the two blue lines represent $CO_2$ concentrations at different heights. The red bars under the $CO_2$ data illustrate the actual occupancy (ground truth), and the yellow bars represent the predicted results, where one indicates occupied, and zero means unoccupied. The baseline scenarios (Fig. 6a/d/g) performed well when the $CO_2$ level rose rapidly due to occupancy. However, as it is solely based on $CO_2$ concentration and its temporal changes, it is often incorrectly predicted as unoccupied states when the room had a relatively low $CO_2$ level due to ventilation, even in cases where the room was occupied (cf. Fig. 6a).

Since the air exchange was mainly through the windows, if occupants closed them instead of ventilating the rooms before leaving, the model would incorrectly predict that the room was still occupied due to the carry-over effect of $CO_2$ (cf. Fig. 6d/g). When utilizing spatial features (cf. Fig. 6b/e/h), the models effectively detected the occupied state even with a low $CO_2$ level due to ventilation. However, since the $CO_2$ in the room requires a certain amount of time to be mixed homogeneously, it can be observed that in some cases (e.g., Fig. 6e), there was a slight prolongation of the predicted occupancy. This issue was optimized by combining temporal and spatial features (Fig. 6f). The slight increase in $CO_2$ concentration around 14:00 in room 1 (Fig. 6a/b/c) might represent an individual case of occupancy not recorded by the questionnaire. The few fluctuations in Fig. 4 when unoccupied are likely due to the same reason.

The contribution of each feature is shown in Fig. 7 using a permutation feature importance method [31]. It is evident that the spatial features of $CO_2$ made the most significant contribution when the combined features were employed, whereas the temporal features played a complementary role in occupancy state detection.

Since physical model-based detection methods often require ventilation information as an input variable, we incorporated the teachers' ventilation rating as a feature to assess the performance of the state classifier. As shown in Table 7, the performance significantly improved by 21.11 percentage points, reaching 89.45 %, when adding the ventilation rating to the baseline scenario. This improvement stems from the strong correlation between ventilation behavior and occupancy in naturally ventilated buildings. The window opening state can only be changed when the room is occupied, enabling the model to capture changes in occupancy through the ventilation rating alone.

### 3.3. Occupancy quantity detection

Similar to Section 3.2, we conducted a quantitative and qualitative evaluation of the occupancy quantity models. Table 8 provides a list of the performance of all models using different $CO_2$-related feature combinations in each room and average values across the three rooms.

Similar to Table 6, the features are categorized into the same three groups: 1) temporal features, 2) spatial features, and 3) combined features. The results in Table 8 have been arranged in ascending order based on the average accuracy of each model.

The relatively higher accuracy and lower RMSE of the occupancy counts in Room 2 compared to the other two rooms may be attributed, in part, to the lower occupancy ratio in Rooms 1 and 3. Additionally, the presence of some occupancy instances not recorded in the questionnaire, potentially affecting the model training, could contribute to the observed differences.

Fig. 8 compares the performance of the occupancy quantity classifiers using average accuracy and RMSE. The utilization of spatial feature VD led to an improvement in accuracy by 8.88 percentage points, reaching 39.55 % compared to the baseline. Although this improvement is not significant, it is noteworthy that there is a substantial increase in prediction accuracy when focusing on room 2 alone (an increase of 35.33 percentage points to 64.00 %, with a standard deviation of 0.06). This observation indicates that quantity classifiers using spatial features may be more sensitive to the dataset than occupancy state classifiers. The combination of temporal and spatial features could enhance the robustness and performance of the model. It is important to point out, as described in Section 3.2, that due to the absence of data using the horizontal difference feature in room 1, the average accuracy of the two best-performing scenarios might be slightly overestimated, and their RMSE potentially underestimated.

In the best-performing scenario (AVG + FD + VD + HD), the accuracy could be substantially improved by a 25.33 percentage points increase, reaching 56.00 %. Simultaneously, the root mean square error (RMSE) exhibited a significant reduction, decreasing by 15 % to achieve a value of 11.44 occupants (normalized RMSE 0.39). This scenario represents a marked enhancement in the model's performance for occupancy quantity detection.

We selected the same days described in Section 3.2 to qualitatively evaluate the best scenarios for each room, as shown in Fig. 9. The black and blue lines illustrate the room's $CO_2$ concentration. The red lines indicate the ground truth of occupancy quantity, and the yellow lines represent the predicted quantity. The baseline scenario (Fig. 9a/d/g) performed poorly due to flaws in the state detection, as explained in Section 3.2. The predicted quantities exhibited frequent fluctuations, and an accuracy of 30.67 % indicates unreliable predictions.

Through the inclusion of spatial features of $CO_2$, the models have demonstrated enhancements in both performance and the reduction of data fluctuations. While some residual noise is still discernible in the prediction results for rooms 1 and 3 (Fig. 9b/c/h/i), there has been a clear improvement compared to the baseline. All scenarios incorporating spatial features exhibited lower RMSE values compared to the baseline.

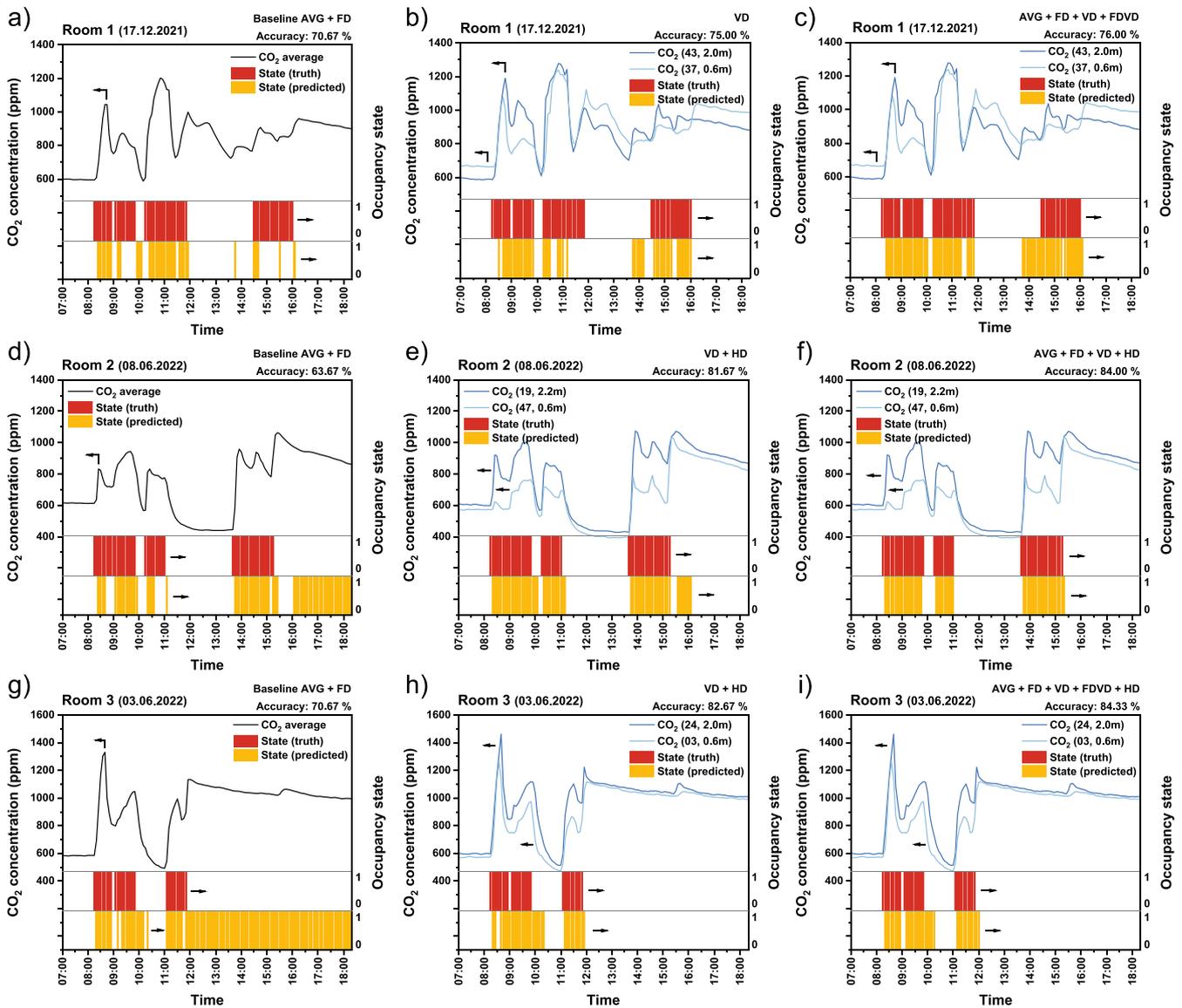

**Fig. 6.** Prediction results of the best scenarios for occupancy state detection in each room under the three feature categories. The lines represent the $CO_2$ concentration in the room. The red bars indicate the ground truth collected through the questionnaire, and the yellow bars are the room state predicted, where one stands for occupied and zero for unoccupied.

Nevertheless, even in the best-performing scenario (Fig. 9f), there were cases of overestimating the occupancy quantity. This reflects that the vertical gradient of $CO_2$ is not strongly correlated with the quantity, as in occupancy state detection, which has a decisive impact.

Although the accuracy of the best scenario was not exceptionally high, the spatial features were still able to improve the classifier's performance compared to the baseline. These results are in line with performance reported by Calí et al. [14] (average accuracy 65.9 %, scenario 3, without window opening information). Furthermore, the average RMSE of the top three scenarios is lower than in related studies (see Table 1). The significance of the features (Fig. 10) also highlights the crucial role played by the spatial features of $CO_2$ in improving $CO_2$-based occupancy quantity detection.

The impact of ventilation information on the model performance in terms of occupancy quantity has also been evaluated in this study, as described in Section 3.2. Table 9 presents the classifier's performance with the inclusion of ventilation rating as a feature. The performance exhibited a modest improvement, increasing by 5.78 percentage points to reach 61.78 %, compared to the best scenario without ventilation information.

## 4. Discussion

### 4.1. Data collection and preprocessing

In comparison with the literature, our dataset contains 4324–4341 data points in the same order of magnitude as the dataset of Zuraimi et al. [15] (6189 data points) when considering the adoption of the same sampling rate. In the baseline scenario, room 2 has slightly lower performance than the other rooms (see Table 6) for both state and quantity detection. One of the reasons could be that as the number of sensors increases, sensors in areas that are further away from the occupants (e.g. sensors 08 & 22) or that are more influenced by the outdoors (e.g. sensor 09) negatively affect the correlation of average $CO_2$ concentration with occupancy. Another reason is possibly that more sensors introduce higher measurement errors from the sensor itself.

For reasons of privacy and user willingness, we were not able to install cameras in the classrooms to monitor the occupancy information. This negatively affects the accuracy of the ground truth labels of our dataset, which is a limitation of this study. Nonetheless, if considering application scenarios, even if data privacy is warranted and communicated to the occupants, the willingness is still a big challenge for the school environments in which our study was conducted. Regarding parental consent, camera-based systems are very difficult to realize in these settings. Therefore, our approach might serve as an alternative for all those situations where camera installation is not desired or feasible. In addition, our proposed features improve the detection performance well when using the same dataset in comparison with the baseline. However, this approach still needs to be validated repeatedly using different datasets. In fact, this is a gap in the literature and a challenge for future work, since most of the published datasets (e.g. [34,35]) do not take into account the heterogeneity of the distribution within the room due to the local $CO_2$ diffusion processes, and only single point measurements are performed in each room. Furthermore, as the original purpose of the installed $CO_2$ sensors was to assess air quality, they were not installed on the ceiling, which is some distance away from the occupants' breathing area. The lack of $CO_2$ data for this area could be another potential limitation.

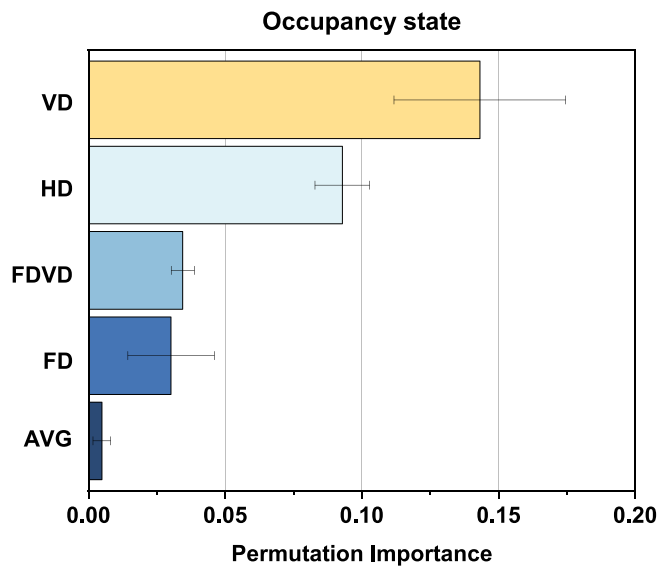

**Fig. 7.** Permutation feature importance for occupancy state detection, average of scenario AVG + FD + VD + FDVD + HD in room 2. Error bars indicate standard deviation.

Calí et al. [14] reported that smoothing the data can improve the performance of the physical model. To ensure comparability with the results of Zuraimi et al. [15] we didn't smooth the data, but we analyzed the effect of the sampling rate on model performance for selected scenarios in room 1. Detailed results are presented in Table A1 in the Appendix. Larger sampling intervals slightly improved the performance because downsampling smoothed the data to some extent and reduced fluctuations in the sensor data. This is consistent with results reported by Calí et al. [14].

### 4.2. Occupancy state detection

When applying combined $CO_2$-related features and ventilation rating while detecting room state, the accuracy scores are generally comparable, with predictions for rooms 1 and 3 revealing slightly lower performance than the baseline scenario with ventilation rating added (2.84 percentage points lower on average). This difference is likely influenced by the impact of the distinct dataset described in Section 2. Despite this, the average accuracy of the two scenarios (88.51 %) is roughly comparable to that reported for naturally ventilated rooms by Calí et al. [14] (72.7 % – 92.3 %, average 84.53 %) using the mass balance equation.

While ventilation information can indeed enhance detection performance significantly, it's essential to note that its effectiveness may vary in mechanically ventilated buildings. Additionally, it carries a potential cost disadvantage, as it necessitates more window sensors.

In the context of detecting room states, adding just one sensor to capture the vertical difference feature brings the performance close to the best scenario (Accuracy: 80.67 %). Detecting horizontal differences requires an additional sensor adjacent to the window, resulting in only a minor performance improvement (Accuracy: 83.17 %). Meanwhile, determining the opening state of the window tends to have a cost disadvantage despite the performance improvement (Accuracy: 89.45 %).

Moreover, even if the opening status of the windows is known, accurately calculating the actual air flow rate is challenging without wind speed information. Various factors, such as air pressure and temperature differences between indoors and outdoors, can affect air exchange markedly, thereby posing challenges for methods based on physical models. Therefore, based on the findings of this study, using two $CO_2$ sensors at different heights would be the optimal low-cost solution for occupancy state detection in naturally ventilated buildings with high performance.

### 4.3. Occupancy quantity detection

The baseline performance of quantity detection falls below the reported performance by Zuraimi et al. [15] using the same $CO_2$-related features. This discrepancy could be attributed, on the one hand, to the more complex ventilation behavior in naturally ventilated buildings and, on the other hand, to differences in room types and user profiles. The occupancy data used by Zuraimi et al. [15] were derived from a large lecture theatre (876 $m^3$, with occupancy exceeding 120 persons, presumably adults.) with mechanical ventilation. In contrast, our data originates from three classrooms in a public school with room volume ranging from 162 $m^3$ to 202 $m^3$, an average occupancy of 19, and a majority of them being children. This resulted in even less significant

**Table 7**
Performance of occupancy state classifiers (average and standard deviation of three rounds) using ventilation rating as an additional feature.

| Feature | Room | Accuracy* | Average Accuracy* | Precision* | Recall* | F1-Score* | Average F1-Score* |
|---|---|---|---|---|---|---|---|
| **Temporal $CO_2$-related Features (baseline) + Ventilation Rating:** | | | | | | | |
| AVG + FD + VENT | 1 | 89.67 % (0.01) | 89.45 % (0.02) | 0.92 (0.00) | 0.90 (0.01) | 0.90 (0.01) | 0.90 (0.02) |
| | 2 | 91.67 % (0.05) | | 0.95 (0.03) | 0.92 (0.05) | 0.93 (0.05) | |
| | 3 | 87.00 % (0.00) | | 0.90 (0.02) | 0.87 (0.00) | 0.88 (0.00) | |
| **Combined $CO_2$-related Features + Ventilation Rating:** | | | | | | | |
| AVG + FD + VD + FDVD + VENT | 1 | 86.00 % (0.04) | 87.56 % (0.03) | 0.92 (0.01) | 0.86 (0.04) | 0.88 (0.03) | 0.89 (0.03) |
| | 2 | 91.67 % (0.05) | | 0.95 (0.03) | 0.92 (0.05) | 0.93 (0.05) | |
| | 3 | 85.00 % (0.01) | | 0.89 (0.02) | 0.85 (0.01) | 0.86 (0.02) | |

* Values in parentheses indicate standard deviations. For an explanation of feature abbreviations, please refer to Table 4.

**Table 8**
Performance of occupancy quantity classifiers (average and standard deviation of three rounds) using only CO$_2$-related features.

| Feature | Room | Accuracy* | Average Accuracy* | RMSE / NRMSE | Average RMSE / NRMSE | F1-Score* | Average F1-Score* |
|---|---|---|---|---|---|---|---|
| **Temporal Features (baseline):** | | | | | | | |
| AVG + FD | 1 | 34.67 % (0.03) | 30.67 % (0.03) | 10.35 / 0.54 | 13.51 / 0.52 | 0.48 (0.02) | 0.43 (0.03) |
| | 2 | 28.67 % (0.04) | | 16.56 / 0.52 | | 0.41 (0.05) | |
| | 3 | 28.67 % (0.02) | | 13.62 / 0.50 | | 0.41 (0.03) | |
| **Spatial Features:** | | | | | | | |
| VD | 1 | 25.33 % (0.10) | 39.55 % (0.17) | 9.23 / 0.49 | 11.01 / 0.44 | 0.37 (0.11) | 0.48 (0.15) |
| | 2 | 64.00 % (0.06) | | 9.90 / 0.31 | | 0.69 (0.05) | |
| | 3 | 29.33 % (0.21) | | 13.89 / 0.51 | | 0.38 (0.21) | |
| VD + HD | 1 | n.a. | 54.34 % (0.11) | n.a. | 11.58 / 0.40 | n.a. | 0.62 (0.10) |
| | 2 | 65.67 % (0.03) | | 10.11 / 0.32 | | 0.71 (0.04) | |
| | 3 | 43.00 % (0.18) | | 13.04 / 0.48 | | 0.52 (0.13) | |
| **Combined Features:** | | | | | | | |
| FD + VD | 1 | 36.33 % (0.23) | 41.22 % (0.15) | 9.58 / 0.50 | 11.46 / 0.45 | 0.46 (0.20) | 0.49 (0.15) |
| | 2 | 61.67 % (0.08) | | 10.41 / 0.33 | | 0.68 (0.06) | |
| | 3 | 25.67 % (0.26) | | 14.39 / 0.53 | | 0.32 (0.26) | |
| AVG + VD | 1 | 38.00 % (0.08) | 41.33 % (0.11) | 8.51 / 0.45 | 11.41 / 0.45 | 0.51 (0.07) | 0.51 (0.11) |
| | 2 | 55.67 % (0.02) | | 12.00 / 0.38 | | 0.64 (0.01) | |
| | 3 | 30.33 % (0.24) | | 13.72 / 0.51 | | 0.37 (0.27) | |
| VD + FDVD | 1 | 29.00 % (0.16) | 42.67 % (0.15) | 8.42 / 0.44 | 10.67 / 0.42 | 0.40 (0.17) | 0.50 (0.13) |
| | 2 | 63.00 % (0.07) | | 10.03 / 0.31 | | 0.69 (0.06) | |
| | 3 | 36.00 % (0.26) | | 13.57 / 0.50 | | 0.42 (0.25) | |
| AVG + FD + VD | 1 | 46.67 % (0.12) | 49.22 % (0.07) | 8.56 / 0.45 | 10.79 / 0.42 | 0.58 (0.10) | 0.60 (0.05) |
| | 2 | 59.00 % (0.03) | | 11.68 / 0.37 | | 0.67 (0.03) | |
| | 3 | 42.00 % (0.11) | | 12.13 / 0.45 | | 0.54 (0.08) | |
| AVG + VD + FDVD | 1 | 41.00 % (0.07) | 50.55 % (0.08) | 8.21 / 0.43 | 10.27 / 0.40 | 0.54 (0.06) | 0.60 (0.05) |
| | 2 | 60.33 % (0.05) | | 11.27 / 0.35 | | 0.67 (0.04) | |
| | 3 | 50.33 % (0.15) | | 11.32 / 0.42 | | 0.59 (0.10) | |
| FD + VD + HD | 1 | n.a. | 52.00 % (0.12) | n.a. | 12.01 / 0.42 | n.a. | 0.59 (0.12) |
| | 2 | 64.33 % (0.06) | | 10.24 / 0.32 | | 0.71 (0.05) | |
| | 3 | 39.67 % (0.23) | | 13.77 / 0.51 | | 0.47 (0.21) | |
| AVG + FD + VD + FDVD | 1 | 47.33 % (0.11) | 53.11 % (0.05) | 8.08 / 0.43 | 10.02 / 0.39 | 0.59 (0.09) | 0.63 (0.03) |
| | 2 | 58.67 % (0.06) | | 11.26 / 0.35 | | 0.67 (0.04) | |
| | 3 | 53.33 % (0.10) | | 10.73 / 0.40 | | 0.63 (0.08) | |
| AVG + FD + VD + FDVD + HD | 1 | n.a. | 56.00 % (0.09) | n.a. | 11.44 / 0.40 | n.a. | 0.64 (0.06) |
| | 2 | 64.67 % (0.06) | | 10.47 / 0.33 | | 0.70 (0.04) | |
| | 3 | 47.33 % (0.13) | | 12.40 / 0.46 | | 0.58 (0.09) | |
| AVG + FD + VD + HD | 1 | n.a. | 56.00 % (0.06) | n.a. | 11.44 / 0.39 | n.a. | 0.65 (0.04) |
| | 2 | 61.67 % (0.06) | | 11.01 / 0.34 | | 0.68 (0.04) | |
| | 3 | 50.33 % (0.06) | | 11.86 / 0.44 | | 0.61 (0.04) | |

* Values in parentheses indicate standard deviations. n.a.: not available due to malfunction of the measurement device. The unit of RMSE is occupancy count. Normalized RMSE (NRMSE) is unitless. Results are listed in ascending order of average accuracy. For an explanation of feature abbreviations, please refer to Table 4.

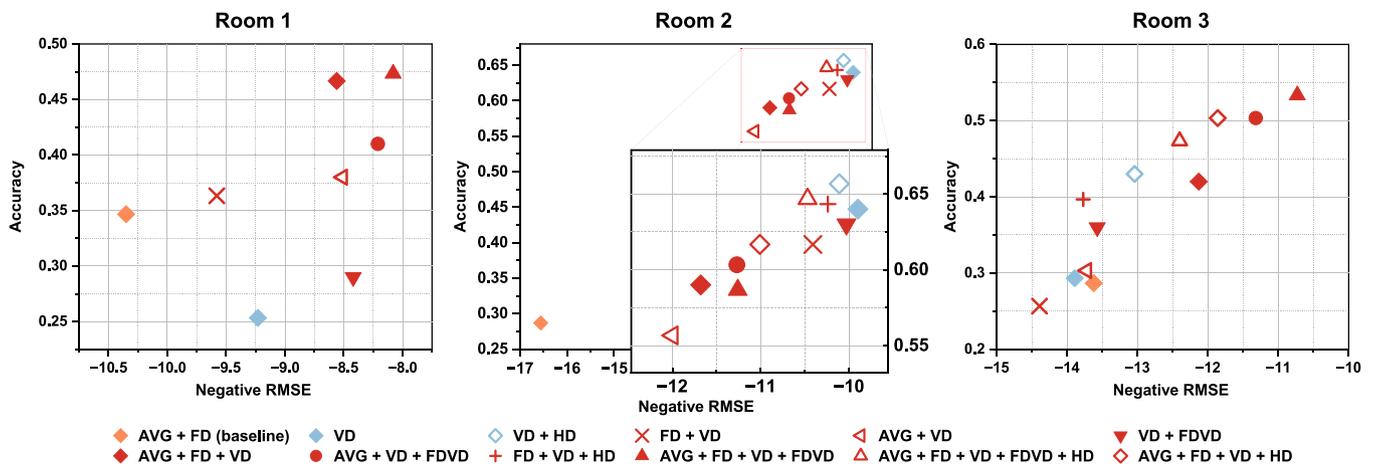

**Fig. 8.** Visualization of quantitative comparisons of quantity classifier performance. Orange points represent classifier performance using temporal features as the baseline, blue for spatial features, and red for combined features. For an explanation of feature abbreviations, please refer to Table 4.

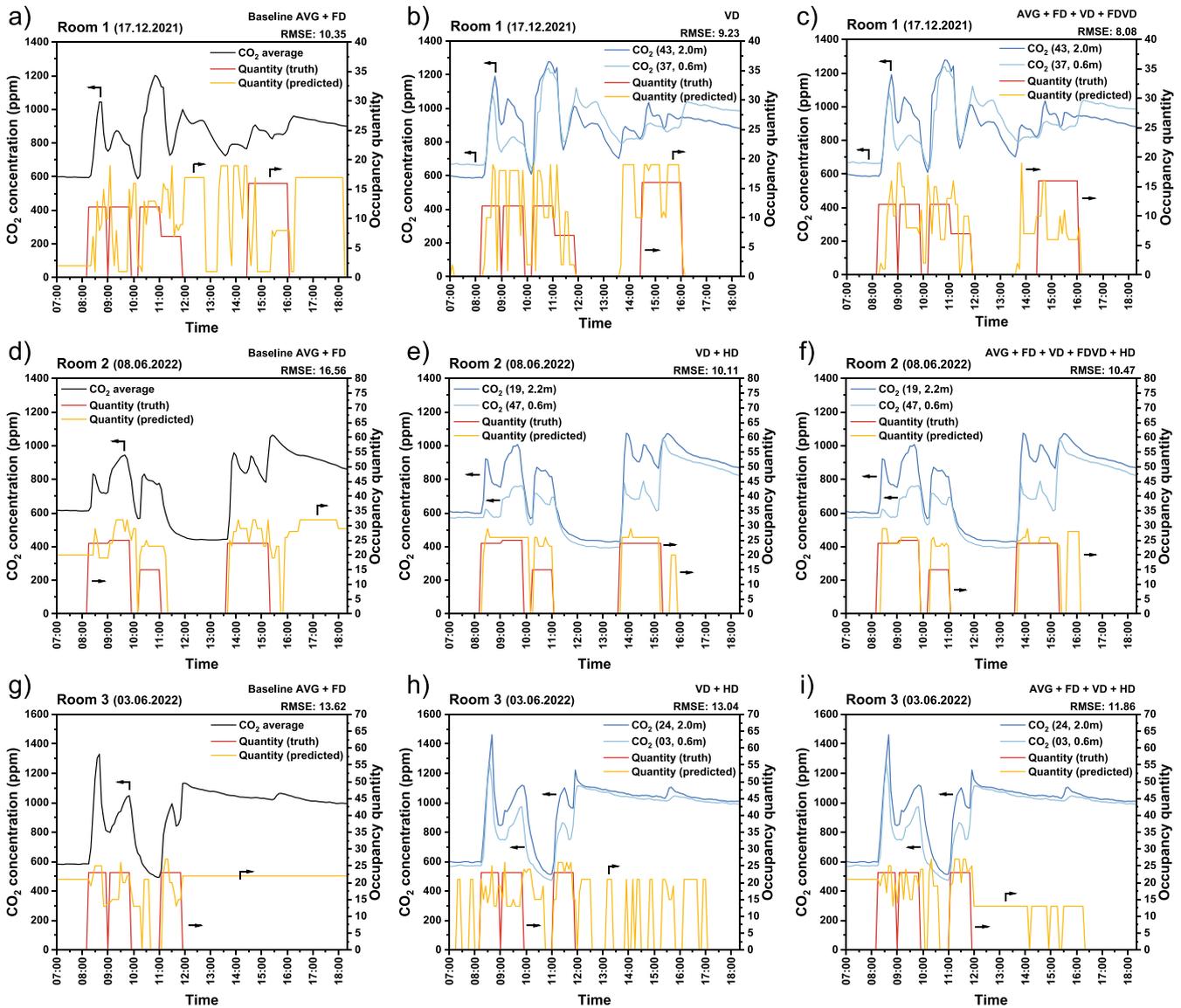

**Fig. 9.** Prediction results of the best scenarios for occupancy quantity detection in each room under the three feature categories. The black and blue lines represent the room's $CO_2$ concentration. The red lines indicate the ground truth of occupancy quantity collected through the questionnaire, and the yellow lines are the predicted quantity.

differences in the temporal features. Additionally, the precision of the data collected through the questionnaire may play a role in this.

The noisy occupancy estimates can be partially attributed to data being considered as temporally independent points by SVMs, which contradicts the strong temporal correlation of $CO_2$. Similar fluctuations in predictions when using SVMs have been reported in related studies [15,16,36]. It is pertinent to mention that the optimization of SVM methods was not a primary focus of this study, as the specific investigation into SVM optimization was not within the scope of our research. Our principal objective centered on comparative feature performance analysis, employing a consistent and proven methodology.

The spatial features could enhance prediction performance. However, even in the best-performing scenario (Fig. 9f), instances of overestimating the occupancy quantity were noted. This observation emphasizes a key aspect: the vertical gradient of $CO_2$ doesn't demonstrate a strong correlation with the quantity, contrasting with its more pronounced and decisive role in occupancy state detection. This highlights the intricacies in accurately predicting occupancy quantity, where factors beyond spatial features may play a more significant role.

The addition of ventilation information only marginally enhances the performance of quantity detection. This is probably due to the inherent difficulty in precisely measuring the actual air exchange in naturally ventilated rooms. As a qualitative assessment, our ventilation rating can only represent the real air exchange to a certain extent.

In comparison to the performance reported by Calí et al. [14] using the mass balance equation, our model's performance (ranging from 59.11 % to 61.78 %, with an average of 60.45 %) closely aligns with their results (scenario 2 with window opening information: ranging from 45.8 % to 80.1 %, with an average of 64.93 %). It's essential to note that our ventilation rating was determined based on the open status of the windows and has not been optimized according to the occupancy quantity data, as in the method employed by Calí et al. [14]. This difference could explain the minor discrepancy in average accuracy.

Summarizing these results, the incorporation of ventilation information did not significantly enhance occupancy quantity detection. The performance is marginally acceptable, regardless of whether ventilation information is included or not. This presents a more promising optimization solution compared to the high cost associated with installing sensors for each window. Potential further performance improvement might be achievable by introducing sensors from other domains, such as humidity, sound, and lighting (cf. [13,16,27,36]). Another challenge is to increase the willingness of users to accept camera-based solutions, which have high accuracy for occupancy detection. These avenues were not further evaluated yet as this study focused on optimizing the use of $CO_2$-related features for occupancy detection.

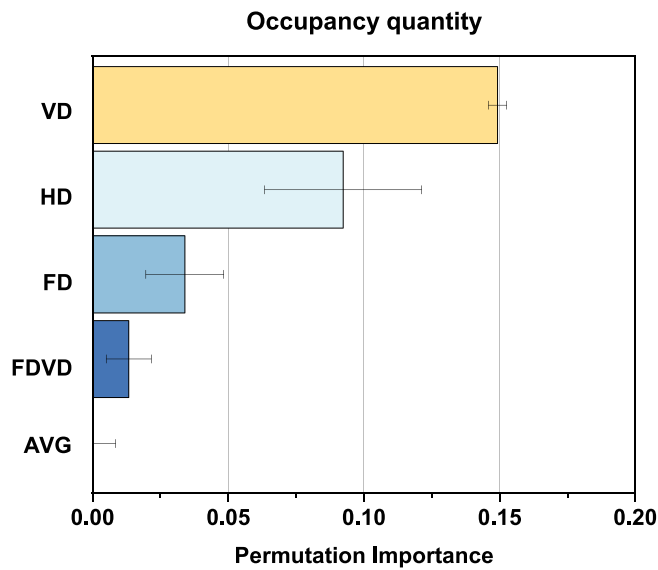

**Fig. 10.** Permutation feature importance for occupancy quantity detection, average of scenario AVG + FD + VD + FDVD + HD in room 2. Error bars indicate standard deviation.

Considering the varying number of $CO_2$ sensors required for different scenarios, Table 10 provides a reference for the performance under each condition to guide the adoption of different sensor options and/or positions. It should be noted that in order to ensure comparability with the results of Zuraimi et al. [15], the models were trained using average $CO_2$ values in the single sensor scenario. In this scenario, the performance of the model may be influenced by the location of the sensor.

## 5. Conclusion

In this study, we employed SVMs for a comprehensive quantitative and qualitative evaluation of $CO_2$-related features. Our introduced spatial feature based on the vertical gradient phenomenon of $CO_2$ demonstrated a substantial enhancement in the performance of $CO_2$-based occupancy detection. Building upon this, the model's performance was further optimized by incorporating the horizontal differences of $CO_2$. When operating within budget constraints, employing one additional sensor markedly improved occupancy detection in naturally ventilated buildings.

The significance of this study lies in introducing novel spatial features grounded in the physical properties of $CO_2$, offering broad applicability to $CO_2$-based occupancy detection methods. The methodology presented in this paper has enhanced occupancy detection performance in naturally ventilated buildings. Future work should prioritize validating the methodology in repeated field studies using more precise ground truth labels, e.g., via camera when feasible. Secondary efforts should focus on refining the methodology to minimize fluctuations in quantity detection and validate its applicability, particularly across various room types and user profiles. With further collection of outdoor climate data and computational fluid dynamics (CFD) simulation, it would assist in the analysis of the $CO_2$ distribution features. This would also help to better understand the spatial heterogeneity of indoor $CO_2$ distribution. In addition, attempts to integrate spatial distribution properties into physical models and investigate the performance of spatial features in buildings with mechanical ventilation are essential directions for further research.

**Table 9**
Performance of occupancy quantity classifiers (average and standard deviation of three rounds) using ventilation rating as an additional feature.

| Feature | Room | Accuracy[*] | Average Accuracy[*] | RMSE[*] | Average RMSE[*] | F1-Score[*] | Average F1-Score[*] |
|---|---|---|---|---|---|---|---|
| **Temporal Features (baseline) + Ventilation Rating:** | | | | | | | |
| AVG + FD + VENT | 1 | 49.67 % (0.05) | 59.11 % (0.07) | 9.13 (0.86) | 9.53 (0.30) | 0.62 (0.03) | 0.67 (0.04) |
| | 2 | 66.33 % (0.06) | | 9.64 (0.30) | | 0.71 (0.05) | |
| | 3 | 61.33 % (0.13) | | 9.83 (1.48) | | 0.69 (0.09) | |
| **Combined Features + Ventilation Rating:** | | | | | | | |
| AVG + FD + VD + FDVD + VENT | 1 | 59.33 % (0.08) | 61.78 % (0.07) | 8.26 (1.39) | 9.02 (1.76) | 0.69 (0.05) | 0.70 (0.05) |
| | 2 | 71.67 % (0.08) | | 7.35 (2.31) | | 0.76 (0.05) | |
| | 3 | 54.33 % (0.09) | | 11.45 (1.72) | | 0.65 (0.06) | |

[*] Values in parentheses indicate standard deviations. The unit of RMSE is occupancy count. For an explanation of feature abbreviations, please refer to Table 4.

**Table 10**
The best performance (average performance across three rooms) in occupancy detection varies with the number of sensors used.

| # | Feature | State | Quantity | |
|---|---|---|---|---|
| | | Accuracy | Accuracy | RMSE[*] |
| 1 | AVG + FD | 68.34 % | 30.67 % | 13.51 |
| 2 | AVG + FD + VD | 80.67 % | 49.22 % | 10.79 |
| 2[**] | AVG + FD + VD + FDVD | 79.56 % | 53.11 % | 10.02 |
| 3 | AVG + FD + VD + FDVD + HD | 83.17 % | 56.00 % | 11.44 |
| 1 + N | AVG + FD + VENT | 89.45 % | 59.11 % | 9.53 |
| 3 + N | AVG + FD + VD + FDVD + VENT | 87.56 % | 61.78 % | 9.02 |

Note. #: number of $CO_2$ sensor.
[*] The unit of RMSE is occupancy count. For an explanation of feature abbreviations, please refer to Table 4.
[**] alternative option, N: multiple window sensors, depending on the number of windows.

**CRediT authorship contribution statement**

**Christoph van Treeck:** Writing – review & editing, Supervision, Funding acquisition. **Jérôme Frisch:** Writing – review & editing, Supervision, Resources, Project administration, Funding acquisition. **Marc Syndicus:** Writing – review & editing, Supervision, Resources, Project administration, Investigation, Funding acquisition. **Qirui Huang:** Writing – original draft, Visualization, Validation, Software, Methodology, Investigation, Formal analysis, Data curation, Conceptualization.

**Declaration of Competing Interest**

The authors declare that they have no known competing financial interests or personal relationships that could have appeared to influence the work reported in this paper.

**Acknowledgement**

The authors thank the pupils, teachers, and school janitors for the possibility of measuring under real conditions.

## Appendix A

Table A1

Model performance to detect occupancy quantity using different sampling rates for data preprocessing for selected scenarios in room 1.

| Sampling rate | Features | Accuracy* | RMSE* | F1-Score* |
|---|---|---|---|---|
| 1 min | AVG + FD | 32.67 % | 10.33 | 0.47 |
| 5 min | AVG + FD | 34.67 % | 10.35 | 0.48 |
| 1 min | VD | 24.00 % | 9.40 | 0.36 |
| 5 min | VD | 25.33 % | 9.23 | 0.37 |
| 1 min | AVG + FD + VD | 45.00 % | 8.19 | 0.57 |
| 5 min | AVG + FD + VD | 46.67 % | 8.56 | 0.58 |

*   Averages across three classrooms. The unit of RMSE is occupancy count. For an explanation of feature abbreviations, please refer to Table 4.

## References

[1] T.A. Nguyen, M. Aiello, Energy intelligent buildings based on user activity: a survey, Energy Build. 56 (2013) 244–257.

[2] Y. Zhang, X. Bai, F.P. Mills, J.C. Pezzey, Rethinking the role of occupant behavior in building energy performance: a review, Energy Build. 172 (2018) 279–294.

[3] L. Rueda, K. Agbossou, A. Cardenas, N. Henao, S. Kelouwani, A comprehensive review of approaches to building occupancy detection, Build. Environ. 180 (2020).

[4] H. Chojer, P. Branco, F. Martins, M. Alvim-Ferraz, S. Sousa, Development of low-cost indoor air quality monitoring devices: recent advancements, Sci. Total Environ. 727 (2020) 138385.

[5] M. Syndicus, Q. Huang, J. Frisch, C. van Treeck, Two-wave intervention study to measure and improve ventilation in classrooms, In: Proceedings of the 18th Healthy Buildings Europe Conference, ISIAQ, Aachen, Germany, 2023.

[6] M. Taştan, A low-cost air quality monitoring system based on Internet of Things for smart homes, J. Ambient Intell. Smart Environ. 14 (2022) 351–374.

[7] J.P. Sá, M.C.M. Alvim-Ferraz, F.G. Martins, S.I. Sousa, Application of the low-cost sensing technology for indoor air quality monitoring: a review, Environ. Technol. Innov. 28 (2022) 102551.

[8] A. Borodinecs, A. Palcikovskis, V. Jacnevs, Indoor Air CO2 sensors and possible uncertainties of measurements: a review and an example of practical measurements, Energies 15 (2022) 6961.

[9] T. Parkinson, A. Parkinson, R. de Dear, Continuous IEQ monitoring system: performance specifications and thermal comfort classification, Build. Environ. 149 (2019) 241–252.

[10] B. Du, M.C. Tandoc, M.L. Mack, J.A. Siegel, Indoor CO 2 concentrations and cognitive function: a critical review, Indoor Air 30 (2020) 1067–1082.

[11] D. Calí, R.K. Andersen, D. Müller, B.W. Olesen, Analysis of occupants' behavior related to the use of windows in German households, Build. Environ. 103 (2016) 54–69.

[12] V. Fabi, R.V. Andersen, S. Corgnati, B.W. Olesen, Occupants' window opening behaviour: a literature review of factors influencing occupant behaviour and models, Build. Environ. 58 (2012) 188–198.

[13] L.M. Candanedo, V. Feldheim, Accurate occupancy detection of an office room from light, temperature, humidity and CO 2 measurements using statistical learning models, Energy Build. 112 (2016) 28–39.

[14] D. Calí, P. Matthes, K. Huchtemann, R. Streblow, D. Müller, CO 2 based occupancy detection algorithm: experimental analysis and validation for office and residential buildings, Build. Environ. 86 (2015) 39–49.

[15] M. Zuraimi, A. Pantazaras, K. Chaturvedi, J. Yang, K. Tham, S. Lee, Predicting occupancy counts using physical and statistical Co2-based modeling methodologies, Build. Environ. 123 (2017) 517–528.

[16] B. Dong, B. Andrews, K.P. Lam, M. Höynck, R. Zhang, Y.-S. Chiou, D. Benitez, An information technology enabled sustainability test-bed (ITEST) for occupancy detection through an environmental sensing network, Energy Build. 42 (2010) 1038–1046.

[17] H. Choi, H. Kim, S. Yeom, T. Hong, K. Jeong, J. Lee, An indoor environmental quality distribution map based on spatial interpolation methods, Build. Environ. 213 (2022) 108880.

[18] G. Pei, D. Rim, S. Schiavon, M. Vannucci, Effect of sensor position on the performance of CO2-based demand controlled ventilation, Energy Build. 202 (2019) 109358.

[19] R.K. Bhagat, M.S. Davies Wykes, S.B. Dalziel, P.F. Linden, Effects of ventilation on the indoor spread of COVID-19, J. Fluid Mech. 903 (2020) F1.

[20] N. Mahyuddin, H. Awbi, The spatial distribution of carbon dioxide in an environmental test chamber, Build. Environ. 45 (2010) 1993–2001.

[21] N. Mahyuddin, H.B. Awbi, M. Alshitawi, The spatial distribution of carbon dioxide in rooms with particular application to classrooms, Indoor Built Environ. 23 (2014) 433–448.

[22] Q. Huang, M. Syndicus, R. Ehrt, A. Sacic, J. Frisch, C. van Treeck, Development of a CO2 based indoor air quality measurement box for classrooms, In: Proceedings of the 18th Healthy Buildings Europe Conference, ISIAQ, Aachen, Germany, 2023.

[23] M. Jin, N. Bekiaris-Liberis, K. Weekly, C.J. Spanos, A.M. Bayen, Occupancy detection via environmental sensing, IEEE Trans. Autom. Sci. Eng. 15 (2018) 443–455.

[24] G.S. Yerragolam, C.J. Howland, R. Yang, R.J. Stevens, R. Verzicco, D. Lohse, Effect of airflow rate on CO2 concentration in downflow indoor ventilation, Indoor Environ. 1 (2024) 100012.

[25] H.E. Beck, N.E. Zimmermann, T.R. McVicar, N. Vergopolan, A. Berg, E.F. Wood, Present and future Köppen-Geiger climate classification maps at 1-km resolution, Sci. Data 5 (2018) 180214.

[26] Sensirion AG, Datasheet Sensirion SCD30 Sensor Module, 2020.〈https://sensirion.com/media/documents/4EAF6AF8/61652C3C/Sensirion_CO2_Sensors_SCD30_Datasheet.pdf〉.

[27] Z. Yang, N. Li, B. Becerik-Gerber, M. Orosz, A systematic approach to occupancy modeling in ambient sensor-rich buildings, Simulation 90 (2014) 960–977.

[28] K.P. Lam, M. Höynck, B. Dong, B. Andrews, Y.-S. Chiou, R. Zhang, D. Benitez, J. Choi, Occupancy detection though an extensive environmental sensor network in an open-plan office building, In: Proceedings of the Eleventh International IBPSA Conference, IBPSA, Glasgow, Scotland, 2009.

[29] S. Wang, J. Burnett, H. Chong, Experimental validation of CO2 -based occupancy detection for demand-controlled ventilation, Indoor Built Environ. 8 (1999) 377–391.

[30] S. Wang, X. Jin, CO2 -based occupancy detection for on-line outdoor air flow control, Indoor Built Environ. 7 (1998) 165–181.

[31] F. Pedregosa, G. Varoquaux, A. Gramfort, V. Michel, B. Thirion, O. Grisel, M. Blondel, P. Prettenhofer, R. Weiss, V. Dubourg, J. Vanderplas, A. Passos, D. Cournapeau, M. Brucher, M. Perrot, E. Duchesnay, Scikit-learn: machine learning in python, J. Mach. Learn. Res. 12 (2011) 2825–2830.

[32] L. Mosley, A balanced approach to the multi-class imbalance problem, Ph.D. thesis, Iowa State University, Ames, Iowa, USA, 2013. 〈https://lib.dr.iastate.edu/etd/13537/〉. 10.31274/etd-180810-3375.

[33] C. Spearman, The proof and measurement of association between two things, Am. J. Psychol. 15 (1904) 72–101.

[34] B. Dong, Y. Liu, W. Mu, Z. Jiang, P. Pandey, T. Hong, B. Olesen, T. Lawrence, Z. O'Neil, C. Andrews, E. Azar, K. Bandurski, R. Bardhan, M. Bavaresco, C. Berger, J. Burry, S. Carlucci, K. Chvatal, M. De Simone, S. Erba, N. Gao, L.T. Graham, C. Grassi, R. Jain, S. Kumar, M. Kjærgaard, S. Korsavi, J. Langevin, Z. Li, A. Lipczynska, A. Mahdavi, J. Malik, M. Marschall, Z. Nagy, L. Neves, W. O'Brien, S. Pan, J.Y. Park, L. Pigliautile, C. Piselli, A.L. Pisello, H.N. Rafsanjani, R.F. Rupp, F. Salim, S. Schiavon, J. Schwee, A. Sonta, M. Touchie, A. Wagner, S. Walsh, Z. Wang, D.M. Webber, D. Yan, P. Zangheri, J. Zhang, X. Zhou, X. Zhou, A global building occupant behavior database, Sci. Data 9 (2022) 369.

[35] S.P. Melgaard, H. Johra, V.Ø. Nyborg, A. Marszal-Pomianowska, R.L. Jensen, K. Kantas, O.K. Larsen, Y. Hu, K.M. Frandsen, T.S. Larsen, K. Svidt, K.H. Andersen,

D. Leiria, M. Schaffer, M. Frandsen, M. Veit, L.F. Ussing, S.M. Lindhard, M.Z. Pomianowski, L. Rohde, A.R. Hansen, P.K. Heiselberg, Detailed operational building data for six office rooms in Denmark: occupancy, indoor environment, heating, ventilation, lighting and room control monitoring with sub-hourly temporal resolution, Data Brief. 54 (2024) 110326.

[36] T. Liu, Y. Li, Z. Bai, J. De, C.V. Le, Z. Lin, S.-H. Lin, G.-B. Huang, D. Cui, Two-stage structured learning approach for stable occupancy detection, In: 2016 International Joint Conference on Neural Networks (IJCNN), IEEE, Vancouver, BC, Canada, 2016, 2306-2312. ⟨http://ieeexplore.ieee.org/document/7727485/⟩. 10.1109/IJCNN. 2016.7727485.

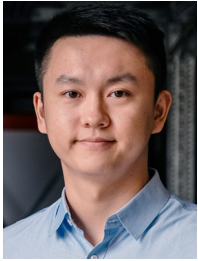

**Qirui Huang**, M.Sc. Research assistant and Ph.D. student at Institute of Energy Efficiency and Sustainable Building (E3D), RWTH Aachen University, Germany.

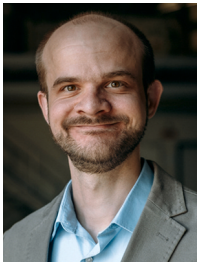

**Marc Syndicus**, Dr. phil. Group leader Ergonomics of the Indoor Environment at Institute of Energy Efficiency and Sustainable Building (E3D), RWTH Aachen University, Germany.

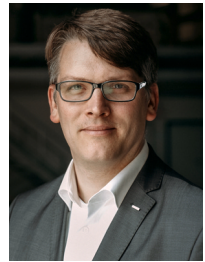

**Jérôme Frisch**, apl. Prof. Dr.-Ing. Extraordinary Professor and Deputy Head of Institute of Energy Efficiency and Sustainable Building (E3D), RWTH Aachen University, Germany.

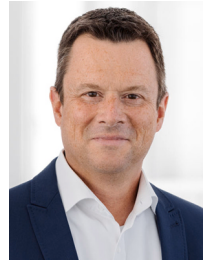

**Christoph van Treeck**, Prof. Dr.-Ing. habil. University Professor and Head of Institute of Energy Efficiency and Sustainable Building (E3D), RWTH Aachen University, Germany.